\documentclass[10pt,twocolumn,letterpaper]{article}

\usepackage{iccv}
\usepackage{times}
\usepackage{epsfig}
\usepackage{graphicx}
\usepackage{amsmath}
\usepackage{amssymb}

\usepackage{mfirstuc}
\usepackage[utf8]{inputenc} 
\usepackage[T1]{fontenc}    
\usepackage{url}            
\usepackage{booktabs}       
\usepackage{amsfonts}       
\usepackage{nicefrac}       
\usepackage{microtype}      
\usepackage{xcolor}         
\usepackage{multirow}
\usepackage{booktabs}
\usepackage[noend]{algpseudocode}
\usepackage{caption}

\usepackage{amsmath,amssymb} 
\usepackage{float}
\usepackage{amssymb}
\usepackage[detect-all]{siunitx}
\usepackage{tikz}
\usepackage{algorithm}
\usepackage{algpseudocode}
\usepackage{graphicx}
\usepackage{subfigure}
\usepackage{amsmath}
\usepackage{dsfont}
\usepackage{color}
\usepackage{ amssymb }
\usepackage{breakcites}
\usepackage{multirow}

\usepackage{tikz}
\usepackage{collcell}
\usepackage{xstring}

\newcommand*{\MaxNumber}{1.0}%
\newcommand*{\MaxNumberL}{30}%
\newcommand*{\MaxNumberJ}{2.5}%
\definecolor{light-gray}{gray}{0.85}

\newcommand{\ApplyGradient}[1]{%
\IfSubStr{#1}{±}{
\StrCut{#1}{±}\csA\csB
\pgfmathsetmacro{\PercentColor}{max(min(100.0*((\csA)/(\MaxNumber)),100.0),0.00)}%
\hspace{-.33em}\colorbox{light-gray!\PercentColor!white}{\framebox[25pt]{#1}}%
}{\IfSubStr{#1}{.}{
\pgfmathsetmacro{\PercentColor}{max(min(100.0*((#1)/(\MaxNumberL)),100.0),0.00)}%
\hspace{-.33em}\colorbox{white!\PercentColor!light-gray}{\framebox[25pt]{#1}}%
}{#1}}}

\newcolumntype{R}{>{\collectcell\ApplyGradient}c<{\endcollectcell}}

\newcommand{\ApplyGradientM}[1]{%
\IfSubStr{#1}{±}{
\StrCut{#1}{±}\csA\csB
\pgfmathsetmacro{\PercentColor}{max(min(100.0*((\csA)/(\MaxNumber)),100.0),0.00)}%
\hspace{-.33em}\colorbox{light-gray!\PercentColor!white}{\framebox[25pt]{#1}}%
}{\IfSubStr{#1}{$\pm$}{
\StrCut{#1}{$\pm$}\csA\csB
\pgfmathsetmacro{\PercentColor}{max(min(100.0*((\csA)/(\MaxNumber)),100.0),0.00)}%
\hspace{-.33em}\colorbox{light-gray!\PercentColor!white}{\framebox[25pt]{\textbf{#1}}}%
}{#1}}}

\newcolumntype{M}{>{\collectcell\ApplyGradientM}c<{\endcollectcell}}

\newcommand{\ApplyGradientL}[1]{%
\IfSubStr{#1}{.}{
\pgfmathsetmacro{\PercentColor}{max(min(100.0*((#1)/(\MaxNumber)),100.0),0.00)}%
\hspace{-.33em}\colorbox{light-gray!\PercentColor!white}{\framebox[10pt]{#1}}%
}{#1}}

\newcolumntype{L}{>{\collectcell\ApplyGradientL}c<{\endcollectcell}}

\newcolumntype{P}[1]{>{\centering\arraybackslash}p{#1}}

\newcommand{\ApplyGradientJ}[1]{%
\IfSubStr{#1}{±}{
\StrCut{#1}{±}\csA\csB
\pgfmathsetmacro{\PercentColor}{max(min(100.0*((\csA)/(\MaxNumberJ)),100.0),0.00)}%
\hspace{-.33em}\colorbox{white!\PercentColor!light-gray}{\framebox[33pt]{#1}}%
}{\IfSubStr{#1}{.}{
\pgfmathsetmacro{\PercentColor}{max(min(100.0*((#1)/(\MaxNumberJ)),100.0),0.00)}%
\hspace{-.33em}\colorbox{white!\PercentColor!light-gray}{\framebox[33pt]{#1}}%
}{#1}}}

\newcolumntype{J}{>{\collectcell\ApplyGradientJ}c<{\endcollectcell}}

\usepackage[breaklinks=true,bookmarks=false]{hyperref}

\iccvfinalcopy 

\def\iccvPaperID{}

\ificcvfinal\fi

\begin{document}

\title{Precise Benchmarking of Explainable AI Attribution Methods}

\author{Rafa\"{e}l Brandt\footnote{Corresponding Author.} \hspace{1.2cm}  Daan Raatjens \hspace{1.2cm} Georgi Gaydadjiev\\
University of Groningen, Groningen, The Netherlands\\
{\tt\small r.brandt@rug.nl}
}

\maketitle

\ificcvfinal\thispagestyle{empty}\fi

\begin{abstract} 
    The rationale behind a deep learning model's output is often difficult to understand by humans. EXplainable AI (XAI) aims at solving this by developing methods that improve interpretability and explainability of machine learning models. Reliable evaluation metrics are needed to assess and compare different XAI methods. We propose a novel evaluation approach for benchmarking state-of-the-art XAI attribution methods. Our proposal consists of a synthetic classification model accompanied by its derived ground truth explanations allowing high precision representation of input nodes contributions. We also propose new high-fidelity metrics to quantify the difference between explanations of the investigated XAI method and those derived from the synthetic model. 
Our metrics allow assessment of explanations in terms of precision and recall separately. In addition, we propose metrics to independently evaluate negative or positive contributions of input nodes. To summarise, our proposal provides deeper insights into XAI methods output.    
    We investigate our proposal by constructing a synthetic convolutional image classification model and benchmarking several widely used XAI attribution methods using our evaluation approach. Moreover, we compare our results with established prior XAI evaluation metrics. 
    By deriving the ground truth directly from the constructed model in our method, we ensure the absence of bias, e.g., subjective either based on the training set. 
Our experimental results provide novel insights into the performance of Guided-Backprop and Smoothgrad XAI methods that are widely in use. Both have good precision and recall scores among positively contributing pixels (0.7, 0.76 and 0.7, 0.77, respectively), but poor precision scores among negatively contributing pixels (0.44, 0.61 and 0.47, 0.75, respectively). The recall scores in the latter case remain close. 
In addition, we show that our metrics are among the fastest in terms of execution time.
\end{abstract}

\section{{Introduction}}
Deep learning, a form of machine learning that uses multi-layered neural networks, has shown groundbreaking results in a variety of fields such as speech recognition, object recognition, genomics, and drug discovery \cite{lecun2015deep}. Deep models are often black boxes, and consequently, the rationale behind their output is frequently difficult to understand. When AI systems make crucial decisions, for example, ones that affect people's lives, it is imperative that all decisions are explainable. To increase the trust that people have in deep models and to improve their accuracy, AI systems should allow humans to understand the rationale behind a deep network's output. Therefore, EU policymakers  demand that AI systems must have the ability to explain their decisions \cite{righttoexplanation}.  

EXplainable Artificial Intelligence (XAI) is a research field that aims at gaining explanations of the internal decision-making process of deep models. However, explanations generated by XAI methods can be misleading and incorrect. It has, for example, been shown in~\cite{adebayo2020sanity} that some XAI methods are independent of both the training data and explained model. Therefore, reliable evaluation metrics are needed to assess and compare different XAI methods. 
The latter is not straightforward since a learned model may contain quirks that cause unexpected behavior. After all, one of the purposes of XAI methods is to discover such oddities. For example, an image object classifier would ideally base its decision only on the pixels that constitute the classified object, but it may have learned to recognize objects by using only parts of it or even the object's background. Ground truth (GT) explanations of trained deep models can therefore be hard to obtain. We propose to alleviate this problem by manually constructing a convolutional image classifier model of which we argue ground truth explanations can be inferred. Furthermore, we propose a set of evaluation metrics that compare GT explanations with those of the XAI evaluated methods. Our implementation is available at \url{https://github.com/rbrandt1/Precise-Benchmarking-of-XAI}.

The main contributions of this paper are as follows:
\begin{itemize}
    \item novel evaluation approach for XAI attribution methods; 
    \item a metric suite to quantify the difference between ground truth and XAI method explanations with high fidelity;
    \item novel metrics to quantify the fidelity of explanations in terms of precision and recall separately, and additional metrics to evaluate negative or positive contributions of input nodes independently, and gain deeper insights into XAI methods output;
    \item careful evaluation of several widely used XAI methods using our evaluation approach, as well as comparison to established prior metrics that measure the fidelity of explanations.
\end{itemize}

The remainder of this paper is organized as follows. Section~\ref{sec:relatedwork} introduces prior explanation fidelity metrics. In Section~\ref{sec:proposed_method}, we propose our evaluation approach for XAI methods. We present the results of the evaluation of several widely used XAI methods in Section~\ref{sec:results} and discuss these in Section~\ref{sec:discussion}. Our experimental setup is detailed in Section~\ref{sec:expsetupt}. Concluding remarks are made in Section~\ref{sec:conclusion}.

\section{{Related Work on XAI Attribution Metrics}}\label{sec:relatedwork}

The sensitivity of an explanation may denote the gradient of the explanation with respect to the input $[\nabla_x\Phi(f(x))]_j = \lim_{\epsilon \rightarrow 0}\frac{\Phi(f(x+\epsilon e_j)) - \Phi(f(x))}{\epsilon}$ \cite{yeh2019infidelity}. Max-Sensitivity \cite{yeh2019infidelity} is a variation of the Sensitivity metric. Given an input $x$, a model $f$, a neighbourhood radius $r$ and an explainer $\Phi$, this metric is defined as $\text{SENS}_{\text{MAX}}(\Phi, f, x, r) = \max_{||y-x|| \leq r} || \Phi (f, y) - \Phi (f, x) ||.$ This equation is used to compute the maximum absolute difference between the two outputs of two slightly different inputs. The second input variation is obtained by adding some noise to the regular input. If this metric scores low, it implies that a small change in the input does not affect the explanation significantly.

Infidelity \cite{yeh2019infidelity} is similar to the sensitivity metric, but instead of adding noise to the input, a large change is performed. A large change to the input should change the output of the model significantly. The infidelity metric is defined as $\text{INFD}(\Phi,f,x) = \mathbb{E}_{I \sim~\mu_i} \Big[ \big( I^T \Phi (f, x) - (f(x) - f(x-I))\big)^2 \Big],$ where $f$ is a model, $\Phi$ is an explanation function, $x$ an input, $I$ is the permutation.

Insertion \cite{Qi_2020} measures the probability change of classification as important pixels are gradually added to an input image.  Deletion \cite{Fong_2017,petsiuk2018rise,alvarezmelis2018robust,vu2020ceval}, measures the decrease in the probability of a class as important pixels from the heatmap are gradually removed from the image. Insertion and Deletion are inspired by the notion that adding noise inside of the irrelevant regions should not affect model's prediction. A variation of Deletion, the remove and retrain (ROAR) metric, ensures the test and train data are of the same distribution by retraining after deletion \cite{hooker2019benchmark}. 

Cosine Similarity is used to measure the similarity between an explanation and ground truth in \cite{GUIDOTTI2021103428}. The metric has range $[0, 1]$, where 1 is the best score.
 
F1-score \cite{GUIDOTTI2021103428} can be used to measure the similarity between an explanation and ground truth. It has range $[0, 1]$, where 1 is the best possible score.

Intersection over union (IoU) is used to measure the similarity between an explanation and ground truth in \cite{DBLP:journals/corr/abs-1712-06302}. It is defined as the number of pixels in the explanation that are truly relevant according to the ground truth divided by size of the union of ground truth and explanation pixels. Its range is $[0, 1]$, where 1 is best. 

Pointing Game \cite{zhang2016topdown} is defined as $\frac{\#hits}{\#hits+\#misses}$, where $\#hits$ is the number of explanations of which the maximum attribution value lies inside a binary GT explanation, and $\#misses$ the number of explanations where this is not the case. Its range is $[0, 1]$, where 1 is best.

Energy-Based Pointing Game \cite{DBLP:journals/corr/abs-1910-01279}, also referred to as The Relevance Mass Accuracy \cite{arras2021ground} of an explanation is a variation of the pointing game that uses the weights of the explanation to calculate a more accurate score. The metric is defined as $ebpg(E,GT) = \dfrac{\sum_{p \in GT \not= 0} E(p)}{\sum_{p \in GT \not= 0} E(p) + \sum_{p \in GT = 0} E(p)}.$ 

The Relevance Rank Accuracy \cite{arras2021ground} of an explanation with respect to a ground truth relevance mask is calculated as the number of pixels in the explanation that are among the top $K$ pixels with the highest relevance that lie within the ground truth relevance mask, divided by $K$. $K$ is the number of pixels in the ground truth relevance mask.

\section{{The Proposed Method}}\label{sec:proposed_method}
We construct a convolutional image classification model that, we argue, can provide reliable ground truth (GT) explanations. Furthermore, we propose a metric suite to quantify the difference between GT and XAI method explanations. 

Our approach is to some degree similar to that of~\cite{arras2021ground}, \cite{DBLP:journals/corr/abs-1712-06302}, \cite{GUIDOTTI2021103428}, \cite{DBLP:journals/corr/abs-2106-12543}, \cite{exploitingpatterns},  \cite{10.1145/3292500.3330930},  \cite{DBLP:journals/corr/abs-2009-02899}, and \cite{zhang2019should}. These works also use (synthetic data and) ground truth to evaluate the performance of XAI methods. However, our approach differs from many of these priors in the way in which the explained model is obtained. Many models in these works are obtained through training. Using training allows the complexity of the explained trained deep models to be more similar to that of typical real-world deep models than our model. However, it has important disadvantages. A common approach is to train a (\eg concept classification) model on a dataset that has concept segmentations available. The concept segmentations are used as GT explanations. Concept segmentations might be inaccurate `GT' explanations as they likely do not depict the difference in discriminating power of different parts (\eg an animal classifier may pay more attention to a face than a tail), and the trained model may over-fit on quirks in the training set (\eg, pay attention to the background of objects) which is not reflected by such `GT' explanations. (Some of) the authors use specially constructed synthetic datasets in an attempt to prevent this from happening. We argue that the construction of a specially crafted training set to combat the above issues, if at all possible, is more complicated than the manual adjustment of the model weights. Therefore, our approach circumvents this problem by carefully constructing a GT model manually rather than obtaining it through training, and by deriving corresponding ground truth explanations from it. The output classes in this dataset are defined by the presence or absence of specific combinations of concepts, just like real-world semantic concepts. 

Note that \cite{exploitingpatterns}, \cite{10.1145/3292500.3330930} and \cite{zhang2019should} also construct classifiers without learning from data (to obtain ground truth explanations), but not CNNs. \cite{GUIDOTTI2021103428} also proposes a way to construct (among others) an image classifier without learning from data, but this classifier is linear while ours is non-linear.

Prior works compare XAI method explanations to ground-truth explanations. In~\cite{DBLP:journals/corr/abs-1712-06302} XAI methods are evaluated with \textit{intersection over union}, \cite{arras2021ground} uses the \textit{Relevance Mass Accuracy} along with \textit{Relevance Rank Accuracy} and \cite{DBLP:journals/corr/abs-2106-12543} uses \textit{infidelity}, \textit{faithfulness}, \textit{monotonocity}, \textit{ROAR} and distance to Shapley values. The evaluation metrics used in \cite{DBLP:journals/corr/abs-2009-02899} are \textit{accuracy}, \textit{recall}, \textit{precision} and \textit{ROC} whereas \cite{GUIDOTTI2021103428} and \cite{10.1145/3292500.3330930} use the \textit{cosine similarity} and \textit{F1-score}. Finally, \cite{exploitingpatterns} uses  the \textit{precision} \textit{recall} and \textit{F1-score} as an evaluation metric.

Our \textit{completeness} and \textit{compactness} metrics are somewhat similar to the conciseness, (energy-based) pointing game, precision, recall, and MAE metrics. Our time metric computes average runtimes of XAI methods similar to \eg, \cite{Qi_2020}.

Prior metrics tend to require binary GT explanations. Ours do not, which is beneficial to their precision. Others that do not, \eg mae, typically do not correct for an imbalance between contributing and not contributing GT pixels unlike ours. Our suite includes metrics that are specifically designed for methods that do not highlight negatively contributing pixels (with a negative value or at all). Also, our metrics allow determining the fidelity of explanations among specific kinds of GT pixels (\eg negatively contributing) separately to allow for detailed analysis. Explanations by XAI methods do not always fall in the [-1, 1] range. Our evaluation approach is compatible with such methods as it includes a normalization step.

\begin{figure*}
    \centering
  
    \scriptsize
     \setlength{\tabcolsep}{0pt}
    \begin{tabular}{P{0.12\textwidth}P{0.12\textwidth}P{0.14\textwidth}P{0.14\textwidth}P{0.14\textwidth}P{0.14\textwidth}P{0.05\textwidth}P{0.05\textwidth}P{0.05\textwidth}P{0.05\textwidth}}
 
          \multicolumn{10}{c}{\includegraphics[width=1\textwidth]{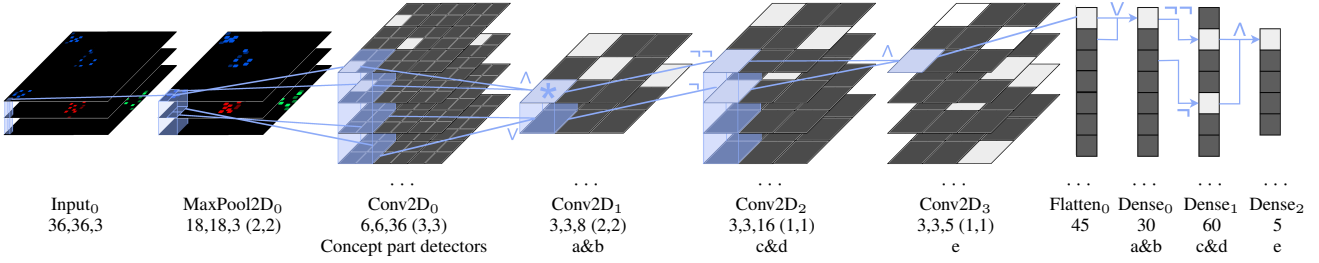}}\\
         & &$\cdots$ &$\cdots$ &$\cdots$ &$\cdots$ &$\cdots$ &$\cdots$ &$\cdots$ & $\cdots$ \\
        Input$_0$ & MaxPool2D$_0$ & Conv2D$_0$ & Conv2D$_1$ & Conv2D$_2$ & Conv2D$_3$ & Flatten$_0$ & Dense$_0$ & Dense$_1$ & Dense$_2$ \\
        36,36,3&18,18,3 (2,2) &6,6,36 (3,3) &3,3,8 (2,2) & 3,3,16 (1,1) & 3,3,5 (1,1) & 45&30&60&5\\
        & &Concept part detectors&a\&b &c\&d&e&&a\&b&c\&d&e 
    \end{tabular}
    \caption{Our GT model.  Examples of elements in previous layer that might influence an element in the next are marked blue. Element $*$ of Conv2D$_1$ might, \eg, compute part \textit{a} of a concept definition, \eg, $(cp_{8,0,0} \wedge cp_{11,1,0})$, and be activated because detector for CP 8 at channel 0 and the one for CP 11 at channel 0 are active at position 0 and 1 respectively. Row 3: Layer output (kernel/stride) size. Row 4: Layer computes this part of a concept/class definition. Conv2D and Dense use ReLU activation clipped between $0$ and $1$, and bias. Valid padding is used.}
    \label{fig:proposedmethod}
\end{figure*}

\subsection{Ground Truth Explanation Model} 
Without loss of generality, we demonstrate our approach using a relatively simple model (Figure~\ref{fig:proposedmethod}), for easy understanding and direct validation, breaking the vicious circle of evaluation approaches being needed to evaluate other evaluation approaches.

\subsubsection{Convolutional Sub-Model}
\begin{figure}
\centering
    \begin{subfigure}
     
    \centering
      \setlength{\tabcolsep}{1pt}
      \renewcommand{\arraystretch}{.75}
\begin{tabular}{ccccc} 
\includegraphics[width=10pt,height=10pt]{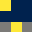}&\includegraphics[width=10pt,height=10pt]{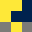}&\includegraphics[width=10pt,height=10pt]{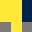}&\includegraphics[width=10pt,height=10pt]{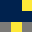}&\includegraphics[width=10pt,height=10pt]{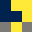}\\
\scriptsize0&\scriptsize1&\scriptsize2&\scriptsize3&\scriptsize4\\
\includegraphics[width=10pt,height=10pt]{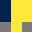}&\includegraphics[width=10pt,height=10pt]{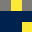}&\includegraphics[width=10pt,height=10pt]{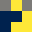}&\includegraphics[width=10pt,height=10pt]{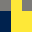}&\includegraphics[width=10pt,height=10pt]{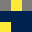}\\
\scriptsize5&\scriptsize6&\scriptsize7&\scriptsize8&\scriptsize9\\
\includegraphics[width=10pt,height=10pt]{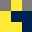}&\includegraphics[width=10pt,height=10pt]{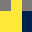} & \multicolumn{3}{c}{ 

\begin{tabular}{p{10pt}p{12pt}p{12pt}}
\multicolumn{3}{c}{\includegraphics[width=30pt,height=5pt]{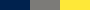}}\\    
\hfill\scriptsize0\hfill&\hfill\scriptsize$\frac{1}{2}$\hfill&\hfill\scriptsize1\hfill
\end{tabular}
  
  }\\
\scriptsize10&\scriptsize11\\
\end{tabular} 
\end{subfigure}\begin{subfigure}
  \scriptsize
    \centering
 \setlength{\tabcolsep}{1pt}
 \renewcommand{\arraystretch}{.75}
\begin{tabular}{ccccc}
\includegraphics[width=10pt,height=10pt]{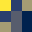}&\includegraphics[width=10pt,height=10pt]{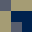}&\includegraphics[width=10pt,height=10pt]{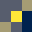}&\includegraphics[width=10pt,height=10pt]{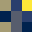}&\includegraphics[width=10pt,height=10pt]{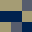}\\
\scriptsize0&\scriptsize1&\scriptsize2&\scriptsize3&\scriptsize4\\
\includegraphics[width=10pt,height=10pt]{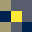}&\includegraphics[width=10pt,height=10pt]{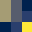}&\includegraphics[width=10pt,height=10pt]{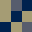}&\includegraphics[width=10pt,height=10pt]{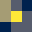}&\includegraphics[width=10pt,height=10pt]{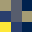}\\
\scriptsize5&\scriptsize6&\scriptsize7&\scriptsize8&\scriptsize9\\
\includegraphics[width=10pt,height=10pt]{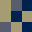}&\includegraphics[width=10pt,height=10pt]{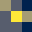} & \multicolumn{3}{c}{ 
\begin{tabular}{p{7pt}p{9pt}p{5pt}p{7pt}p{7pt}}
\multicolumn{5}{c}{\includegraphics[width=30pt,height=5pt]{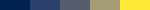}}\\    
\hfill\scriptsize-1\hfill&\hfill\scriptsize-$\frac{1}{2}$\hfill&\hfill\scriptsize0\hfill&\hfill\scriptsize1\hfill&\hfill\scriptsize2\hfill
\end{tabular}
  
  }\\
\scriptsize10& \scriptsize11\\
\end{tabular}
    \end{subfigure}\begin{subfigure}
      \scriptsize
    \centering
        \begin{tabular}{l}
            \includegraphics[width=10pt]{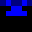}\\
            \includegraphics[width=10pt]{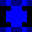}\\
            \includegraphics[width=10pt]{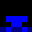}
        \end{tabular}
        \end{subfigure}
    \caption{Concepts (three examples on the right) consist  of concept parts (left) that are detected with weights (middle).}\label{fig:concept_parts}
\end{figure}

The \textit{convolutional sub-model} classifies the concepts contained in input images.  Concepts are parts of RGB input images with a size of $6 \times 6$ pixels. Concept examples can be divided in four non-overlapping parts of $3 \times 3$ pixels. These parts may contain, in each of their channels, one of the concept parts illustrated in Figure~\ref{fig:concept_parts}a, or only zeros. Concepts are defined based on the absence or presence of these parts. All concept definitions follow a syntax that allows for the manual construction of the \textit{Convolutional sub-model}, i.e., without learning from examples. We denote a concept part (CP) $CP\in\bigcup\limits^{11}_{id=0}\bigcup\limits^{3}_{pos=0}\bigcup\limits^{2}_{ch=0}\{cp_{id,pos,ch}\}\cup\{\top\},$ where $id$ denotes concept part identifier as illustrated in Figure~\ref{fig:concept_parts}a and b, $pos$ denotes the  position row-wise from top left to bottom right, and $ch$ denotes channel. Both, concept and class definitions contain unary operators $OP_1\in\{\neg\neg,\neg\}$ (note that $\neg\neg P \leftrightarrow P$), and binary operators $OP_2\in\{\vee,\wedge\}.$  

Concept definitions have syntax $(\underbrace{OP_1}_\text{c} ( CP \underbrace{OP_2}_\text{a} CP))  \underbrace{OP_2}_\text{e} (\underbrace{OP_1}_\text{d} (CP \underbrace{OP_2}_\text{b} CP ))$. 

Concepts recognized by the \textit{convolutional sub-model} are
\begin{equation}
\begin{tabular}{l}
$0: (  ( cp_{7,0,0}  \wedge  cp_{10,1,0} ) ) \vee ( (cp_{4,2,0}  \wedge  cp_{1,3,0}))$\\
$1: (  ( cp_{6,0,0}  \wedge  cp_{9,1,0} ) ) \wedge ( (cp_{3,2,0}  \wedge  cp_{0,3,0}))$\\
$2: (  ( cp_{7,0,1}  \vee  cp_{10,1,1} ) ) \wedge (  (cp_{4,2,2}  \wedge  cp_{1,3,2}))$\\
$3: (  ( cp_{8,0,1}  \wedge  cp_{11,1,1} ) ) \wedge (  (\top  \wedge  \top))$\\
$4: (  ( cp_{8,0,1}  \vee  cp_{11,1,1} ) ) \wedge (\neg (cp_{8,0,1}  \wedge  cp_{11,1,1}))$\\
\end{tabular}
\end{equation}
Examples of concept 0 are shown in Figure~\ref{fig:concept_parts}c.

The \textit{convolutional sub-model} consists of four 2D convolutional layers. The first layer (i.e., Conv2D\_0) detects concept parts $CP$. Concept part detectors have weights as detailed in Figure~\ref{fig:concept_parts}b. The filters have $-1$ bias. Each concept part detector detects a concept part in a specific channel. Layer Conv2D\_1 computes part $a$ and $b$ of the concept definition syntax. $OP_2\vee$ has input weights $\{1,1\}$ and bias $0$. $OP_2\wedge$ has input weights $\{1,1\}$ and bias $-1$. Conv2D\_2 computes part $c$ and $d$.  Each filter has one input node with non-zero weight. $OP_1\neg$ has input weight $-1$ and bias $1$. $OP_1\neg\neg$ has input weight $1$ and bias $0$. Conv2D\_3 computes part $e$. Weights are set in the same way as those of Conv2D\_1. Except, the filter corresponding to concept 3 has a single input node (with weight $1$) and bias $0$.

\subsubsection{Dense Sub-Model}

The \textit{dense sub-model} classifies input examples based on the concepts contained therein as determined by the \textit{convolutional sub-model}. Input examples have size $3 \times 3$ concepts. A class example belongs to a class when a combination of concepts at specific positions are contained or not contained in the input, as defined in a class definition.   A concept at a position is denoted $C\in\bigcup\limits^{4}_{id=0}\bigcup\limits^{8}_{pos=0}\{c_{id,pos}\}\cup\{\top\}.$ Classes have syntax $( \underbrace{OP_1}_\text{c}  ( C \underbrace{OP_2}_\text{a}  C ))  \underbrace{OP_2}_\text{e}  (\underbrace{OP_1}_\text{d}  ( C \underbrace{OP_2}_\text{b}  C )).$ 

Let $id \in [0,4]$ denote a concept. The recognized classes:
\begin{equation}\label{eq:classdefinitions}
\begin{tabular}{ll}
$0:~(( c_{id,0}\wedge\top ) ) \wedge ( (\top\wedge\top))$\\
$1:~(( c_{id,3}\wedge\top ) ) \wedge (\top\wedge\top)$\\
$2:~(( c_{id,1}\wedge c_{id,2} ) ) \wedge ((\top\wedge\top))$\\
$3:~(( c_{id,1}\vee c_{id,2} ) ) \wedge (\neg (c_{id,1}\wedge c_{id,2}))$\\
$4:~( \neg ( c_{id,0}\vee c_{id,1} ) ) \wedge (\neg (c_{id,2}\vee c_{id,3}))$.
\end{tabular}
\end{equation}

The \textit{dense sub-model} consists of three dense layers. The first (i.e., Dense\_0) computes part $a$ and $b$ of the class syntax. In general, $OP_2\vee$ has input weights $\{1,1\}$ and bias $0$, and $OP_2\wedge$ has input weights $\{1,1\}$ and bias $-1$. However, nodes corresponding to one of these terms where one of the arguments is $\top$ have a single input with weight $1$ and bias $0$. Dense\_1 computes part $c$ and $d$.  Each node has one input node with non-zero weight. $OP_1\neg$ has input weight $-1$ and bias $1$. $OP_1\neg\neg$ has input weight $1$ and bias $0$. Dense\_2 computes part $e$, its weights are set like Dense\_0 weights. 

\subsubsection{Ground Truth Explanations}\label{sec:ground-truth}

Sets of pixel-channel elements (a RGB pixel consists of three pixel-channel elements) that may belong to ground-truth explanations include (1) Those that contribute positively (or negatively) to the true class score, (2) Those that  would contribute positively (or negatively) to the true class score if they were different, (3) Those that contribute positively (or negatively) to one or more classes other than the true class, and (4) those that would contribute positively (or negatively) to one or more classes other than the true class if they were different. Our ground truth explanations consist of pixels in set 1 with respectively positive and negative values as these are the pixels that influenced the predicted true class score. This might contradict with the pixel-channel elements an XAI method is designed to highlight or the way in which contribution values are designed to be encoded.

We determine the influence of pixel-channel elements on class scores from the last to the first layer. Both of the $OP_1$ operators and the $OP_2\wedge$ operator feed full influence to all of their input nodes which have non-zero weight. The $OP_2$ operator $\vee$ divides the influence between activated input nodes that have non-zero weight. Note that classes and concepts with an `OR' in their definition use their weight vectors to obtain invariance, a common characteristic of trained ANNs.
Concept part detectors have weights as detailed in Figure~\ref{fig:concept_parts}b. Each concept part has in either the top or bottom row $[\frac{1}{2}, 1, \frac{1}{2}]$ with accompanying weights $[1, 0, -1]$. We expect an explanation of these rows of $[\frac{1}{2}, 0, -\frac{1}{2}]$, except adjusted for backtracked influence scores. Each concept part has in either the left or right column $[0;0]$ with accompanying weights $[1;-1]$. We expect an explanation of these rows of $[0;0]$. We added this noise in our approach to make our ground truth model more similar to trained ANNs. 

The influence score of a concept at a position is multiplied with its corresponding concept part detector weights to obtain a pixel-channel element level ground truth explanation. In case the concept contributes positively to the class score, the explanation is multiplied element wise with the corresponding pixels in the input. In case the concept contributes negatively to the class score, the original explained value is multiplied by (1 - the input value). 

Ground truth and XAI method explanations are individually normalized between -1 and 1. To find the normalized pixel intensity of image $I$ at position $p$, we compute $normalize(I,p) = I(p)/(max(abs(min(I)),abs(max(I)))).$

Ground truth explanations are three-dimensional. Some XAI methods provide a value per ($u,v$) coordinate rather than per ($u,v,w$) coordinate. To make our ground truth explanations compatible with these methods, 2D ground truth explanations are generated. These are equal to 3D variants, except they are converted to 2D by taking the value that is furthest from 0 (positive or negative) over channels. 3D ground truth explanations have the same number of non-zero valued pixel-channel elements as their 2D version and $2 \times width \times height$ additional zero-valued elements.

\subsubsection{Construction of Evaluation Data Set}
Five models and corresponding test sets were constructed to avoid examples belonging to multiple classes simultaneously. Each model has a different concept $id \in [0,4]$ (Equation~\ref{eq:classdefinitions}). The models each have binary cross-entropy loss $0.0000$.  Sixteen (16) examples were constructed for each class. To construct an example of a class, by its definition required concepts are inserted and other concepts are inserted randomly while ensuring the example is an example of a single class. Class definitions stating that concepts containing $\vee$ statements may not be present have ambiguity regarding their ground truth explanation. The test sets therefore do not contain instances of class $3$ and $4$ with concepts $0$, $2$, and $4$.

At most one concept is present at a (x,y) coordinate as it would otherwise be ambiguous whether a 2D  explanation correctly highlighted an influential concept, or mistakenly highlighted an uninfluential concept at the same coordinate.

Let $R_n, n\in[0,1]$ be a random variable following the discrete uniform distribution over the set \{0,1\}. To determine to what extent XAI methods are able to cope with the max-pooling layer, both input examples and ground truth explanations are up-scaled with synchronized $R_n$: $upscaled(y*2+R_0, x*2+R_1, :) \gets img(y, x, :)$ for $y \in [0,18)$, $x \in [0,18)$.

\subsection{Evaluation Metrics}\label{sec:metrics}
\begin{figure}
    \centering
      \scriptsize
      \setlength{\tabcolsep}{0pt}
      
\begin{tabular}{c}
          \begin{tabular}{c}
            \includegraphics[width=15pt]{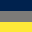}\\
             Explanation
        \end{tabular} \\\\
        \begin{tabular}{c}
            \includegraphics[width=15pt]{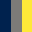}\\
            GT
        \end{tabular} 
\end{tabular}~~~~~
\begin{tabular}{r@{~~}|@{~~}c@{~~}c@{~~}c@{~~}|@{~~}c@{~~}c@{~~}c}
            Metric & \multicolumn{2}{l}{Compactness} && \multicolumn{3}{l}{Completeness}\\
            Metric Variant &$!=$&$>$&$<$&$!=$&$>$&$<$\\
    \begin{tabular}{r}
           Numerator  \\\\
           Denominator
       \end{tabular}  & \begin{tabular}{r@{~}l}
            \includegraphics[width=15pt]{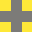} \\
            \includegraphics[width=15pt]{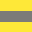}  
       \end{tabular} &
        \begin{tabular}{r@{~}l}
            \includegraphics[width=15pt]{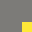} \\
            \includegraphics[width=15pt]{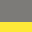}  
       \end{tabular} &
        \begin{tabular}{r@{~}l}
            \includegraphics[width=15pt]{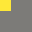} \\
            \includegraphics[width=15pt]{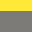}  
       \end{tabular} & 
       \begin{tabular}{r@{~}l}
            \includegraphics[width=15pt]{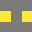} \\
            \includegraphics[width=15pt]{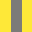}  
       \end{tabular} &
        \begin{tabular}{r@{~}l}
            \includegraphics[width=15pt]{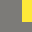} \\
            \includegraphics[width=15pt]{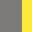}  
       \end{tabular} &
        \begin{tabular}{r@{~}l}
            \includegraphics[width=15pt]{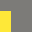} \\
            \includegraphics[width=15pt]{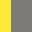}  
       \end{tabular} \\
              Value & $\frac{2}{3}$ & $\frac{1}{3}$& $\frac{1}{3}$& $\frac{2}{3}$ & $\frac{1}{3}$& $\frac{1}{3}$
      
  \end{tabular}~~~~~\begin{tabular}{c}
            -1\\
            \includegraphics[height=30pt,width=5pt]{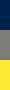}\\
            1
        \end{tabular}
  
    \caption{{Our metrics are computed by dividing the aggregate sums of the top and bottom images. Completeness is one minus the result.}}
    \label{fig:metricvis}
\end{figure}

To measure the accuracy of an explanation $E$ of a prediction made by our ground truth model, we compare the explanation to ground truth explanation $GT$ in terms of \textit{Completeness}, \textit{Compactness}, and \textit{Correctness} as detailed below and illustrated in Figure~\ref{fig:metricvis}. XAI methods may produce a separate explanation per concept. We denote the (ground truth) explanation with respect to concept $c \in C$ as ($GT_c$) $E_c$. In the following, image coordinate $(u,v,w)$ or $(u,v)$ (depending on the dimensions of the explanation) is denoted $p$. Explanations are assumed to be normalized (Section~\ref{sec:ground-truth}). All of our metrics except \textit{Time} have values range $[0,1]$, where 1 is best. \textit{Time} values are in $[0,\infty)$, where 0 is the best value.

\subsubsection{Completeness}
The \textit{completeness} of an explanation $cpl(E, GT, \not=)$ describes the mean absolute error of an explanation among pixels that actually influenced the true class score (positively or negatively). The sign of the (explained) influence of pixels is disregarded. For concept-level explanations $cpl(E,GT,\not=)\stackrel{\text{def}}{=}1-\frac{1}{\#C}\sum_{c \in C} \Psi(E_c,GT_c)$ and for image-level explanations $cpl(E,GT,\not=)\stackrel{\text{def}}{=}1- \Psi(E,GT)$, where $\Psi(E,GT)\stackrel{\text{def}}{=}$ \[\begin{cases}\frac{\sum_{p\in GT\not=0}||E(p)|-|GT(p)||}{\#(GT\not=0)}&\textbf{if}~\#(GT\not=0)\not=0\\ 
     0&\textbf{else}. 
    \end{cases}\] 
    
The \textit{completeness} of an explanation $cpl(E, GT, >)$ ($cpl(E, GT, <)$) describes the mean absolute error of an explanation among pixels that influenced the true class score positively (or negatively). For concept-level explanations $cpl(E, GT, \bullet) \stackrel{\text{def}}{=} 1-\frac{1}{\# C}\sum_{c \in C} \Upsilon(E_c,GT_c) $ and for image-level explanations $cpl(E, GT, \bullet) \stackrel{\text{def}}{=} 1- \Upsilon(E,GT),$ where $\bullet \in \{<, >\}$, $\Upsilon(E,GT)\stackrel{\text{def}}{=}$ \[\begin{cases}
\frac{\sum_{p \in GT \bullet 0} |E(p) - GT(p)|_{[0,1]}}{\#(GT\bullet0)}&\textbf{if}~\#(GT\bullet0)\not=0\\ 
0&\textbf{else},
\end{cases}\]$|x|_{[0,1]}$ is the absolute value of $x$ clipped between $0$ and $1$.

\subsubsection{Compactness}
The \textit{compactness} of an explanation $cpa(E, GT, \not=)$ describes the degree to which the total attribution assigned by a XAI method is made to pixels that actually influenced the true class score. $cpa(E, GT, >)$ ($cpa(E, GT, <)$) describes the compactness of an explanation among pixels that do not contribute positively (negatively) to the true class. $\begin{aligned}
cpa(E, GT, \bullet) \stackrel{\text{def}}{=} 
    \frac{1}{\#C}\sum_{c \in C} \Phi(E_c,GT_c)
\end{aligned}$ (concept-level explanations) or $cpa(E, GT, \bullet) \stackrel{\text{def}}{=} \Phi(E,GT)$ (image-level explanations), where $\Phi(E,GT)\stackrel{\text{def}}{=}$ \[\begin{cases}
    \frac{TP_{acc}(E,GT)}{TP_{acc}(E,GT)+FP_{err}(E,GT)} &\textbf{if}~TP_{acc}\not=0\\ 
     1&\textbf{if}~TP_{acc}=FP_{err}=0\\
     0&\textbf{else},
    \end{cases},\] where $TP_{acc}(E,GT)\stackrel{\text{def}}{=}\\\sum_{p\in(GT\bullet0 \wedge E\bullet0)}(1-f(E(p),GT(p),\bullet)),$\\
$FP_{err}(E,GT)\stackrel{\text{def}}{=}\sum_{p\in(\neg GT\bullet0\wedge E\bullet0)}f(E(p),GT(p),\bullet),$
and $\begin{aligned}
    f(e, gt,\bullet) \stackrel{\text{def}}{=} 
    \begin{cases}
    ||e|-|gt|| & \textbf{ if } \bullet \text{is} \not= \\ 
   |e - gt|_{[0,1]} & \textbf{else} 
    \end{cases}.
\end{aligned}$

\subsubsection{Correctness} 
\textit{Correctness} aggregates \textit{compactness} and \textit{completeness} into a single value, i.e. $cor(E, GT, \bullet) \stackrel{\text{def}}{=} \frac{1}{2}\left(cpl(E, GT, \bullet) + cpa(E, GT, \bullet)\right).$ 
In order to get a sense of the correctness of the explained influence of pixel-channel elements, irrespective of whether the sign of this influence was correctly explained we define $cor_{ns}(E, GT) \stackrel{\text{def}}{=}  cor(E, GT, \not=).$
To quantify the correctness of the explained influence of pixel-channel elements, taking into account whether the sign of this influence was correctly explained we define $cor_{s}(E, GT)~\stackrel{\text{def}}{=}~\frac{1}{2}\left(cor(E, GT, >) + cor(E, GT, <)\right).$

\subsubsection{Time} 

We compute the average computation time of an XAI method in milliseconds per explanation over the test set. 

\section{{Experimental Setup}}\label{sec:expsetupt}
We assessed our evaluation metrics and compared them to the metrics MuFidelity (MF)$^*$ \cite{bhatt2020evaluating}, Deletion (Del)$^*$ and  Insertion (Ins)$^*$ \cite{petsiuk2018rise}, structural similarity index measure (SSIM)$^+$ \cite{hassan2012structural}, Conciseness (CSN) \cite{Amparore_2021}, Cosine Similarity (COS) \cite{GUIDOTTI2021103428},  Relevance Rank Accuracy (RRA) \cite{arras2021ground}, Mean absolute error (MAE)$^T$, and energy-based pointing game (EBPG) \cite{DBLP:journals/corr/abs-1910-01279} (GT explanations were turned binary for EBPG and RRA). Also, Intersection over union (IoU)$^T$, Precision (PR)$^T$, Recall (RE)$^T$, and F1$^T$ \cite{GUIDOTTI2021103428} were included, in these cases GT and XAI explanations were turned binary. To this end, the XAI methods Deconvolution$^*$ and Occlusion$^*$ \cite{DBLP:journals/corr/ZeilerF13}, Grad-CAM$^*$ \cite{gradcam2019}, Grad-CAM++$^*$ \cite{gradcamplus2018}, Gradient Input$^*$ \cite{ancona2018better}, Guided Backprop$^*$ \cite{springenberg2015striving}, Integrated Gradients$^*$ \cite{sundararajan2017axiomatic}, Saliency$^*$ \cite{simonyan2014deep}, SmoothGrad$^*$ \cite{smilkov2017smoothgrad}, SquareGrad$^*$ \cite{hooker2019benchmark},  VarGrad$^*$ \cite{adebayo2020sanity},  Rise$^*$ \cite{petsiuk2018rise}, KernelShap$^*$ \cite{lundberg2017unified}, and Lime$^*$ \cite{ribeiro2016should} have been executed. We use the implementation offered in the \cite{Xplique} library of all XAI methods and metrics marked $^*$. An implementation of the metrics Average Stability, MeGe and ReCo is offered by said library, but they were excluded as they do not measure fidelity or assume the explained model is obtained through training which our model is not.  We use the \textit{skimage}\cite{skiimage} (\textit{skilearn}\cite{skilearn}) implementation of metrics marked $^+$ ($^T$). We normalized explanations  (Sec.~\ref{sec:ground-truth}) and used \cite{keras}.

To reduce the number of warnings raised during MuFidelity computation, and to prevent ``not a number'' output of Cosine Similarity, the pixel at the origin of XAI method explanations was set to $1e-9$ in all channels when the explanation would be all zero otherwise. Be cautioned that MuFidelity warnings were not completely removed. Rise explanations seem offset/biased. The cause is not clear to us. 

Our experiments used an Intel{\small\textregistered~} Core{\small\texttrademark~} i7-2600K CPU @3.40GHz with 8GB ram. A NVIDIA GeForce RTX 2060 GPU was available to methods and metrics in marked cases.

\section{{Experimental Results}}\label{sec:results}

\begin{figure*}[htb]
    \centering
    \scriptsize
    \newcommand{\figwidthvar}{0.05}
\begin{tabular}{c@{~}c@{~}c@{~}c@{~}c@{~}c@{~}c@{~}c@{~}c@{~}c@{~}c@{~}c@{~}c@{~}c@{~}c@{~}c@{~}r}
\includegraphics[width=\figwidthvar\linewidth]{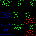} &
 \includegraphics[width=\figwidthvar\linewidth]{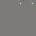} &
 \includegraphics[width=\figwidthvar\linewidth]{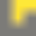} &
 \includegraphics[width=\figwidthvar\linewidth]{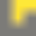} & 
 \includegraphics[width=\figwidthvar\linewidth]{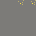} & 
 \includegraphics[width=\figwidthvar\linewidth]{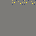} & 
 \includegraphics[width=\figwidthvar\linewidth]{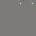} & 
 \includegraphics[width=\figwidthvar\linewidth]{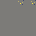} & 
 \includegraphics[width=\figwidthvar\linewidth]{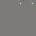} & 
 \includegraphics[width=\figwidthvar\linewidth]{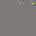} & 
 \includegraphics[width=\figwidthvar\linewidth]{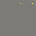} & 
 \includegraphics[width=\figwidthvar\linewidth]{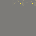} &
  \includegraphics[width=\figwidthvar\linewidth]{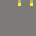} & 
\includegraphics[width=\figwidthvar\linewidth]{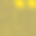} &
\includegraphics[width=\figwidthvar\linewidth]{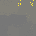} &
\includegraphics[width=\figwidthvar\linewidth]{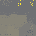} &
  \multirow{5}{*}{\begin{tabular}{c}
  -1\\
  \includegraphics[width=5pt,height=50pt]{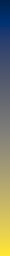}\\
  1
  \end{tabular}}\\
  \includegraphics[width=\figwidthvar\linewidth]{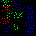} &
 \includegraphics[width=\figwidthvar\linewidth]{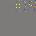} &
 \includegraphics[width=\figwidthvar\linewidth]{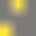} &
 \includegraphics[width=\figwidthvar\linewidth]{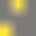} & 
 \includegraphics[width=\figwidthvar\linewidth]{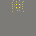} & 
 \includegraphics[width=\figwidthvar\linewidth]{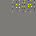} & 
 \includegraphics[width=\figwidthvar\linewidth]{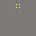} & 
 \includegraphics[width=\figwidthvar\linewidth]{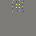} & 
 \includegraphics[width=\figwidthvar\linewidth]{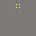} & 
 \includegraphics[width=\figwidthvar\linewidth]{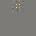} & 
 \includegraphics[width=\figwidthvar\linewidth]{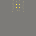} & 
 \includegraphics[width=\figwidthvar\linewidth]{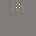}&
  \includegraphics[width=\figwidthvar\linewidth]{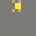} &
  \includegraphics[width=\figwidthvar\linewidth]{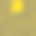} &
\includegraphics[width=\figwidthvar\linewidth]{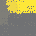} &
\includegraphics[width=\figwidthvar\linewidth]{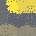}  \\
  \includegraphics[width=\figwidthvar\linewidth]{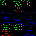} &
 \includegraphics[width=\figwidthvar\linewidth]{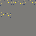} &
 \includegraphics[width=\figwidthvar\linewidth]{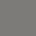} &
 \includegraphics[width=\figwidthvar\linewidth]{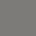} & 
 \includegraphics[width=\figwidthvar\linewidth]{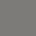} & 
 \includegraphics[width=\figwidthvar\linewidth]{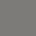} & 
 \includegraphics[width=\figwidthvar\linewidth]{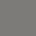} & 
 \includegraphics[width=\figwidthvar\linewidth]{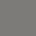} & 
 \includegraphics[width=\figwidthvar\linewidth]{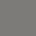} & 
 \includegraphics[width=\figwidthvar\linewidth]{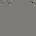} & 
 \includegraphics[width=\figwidthvar\linewidth]{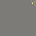} & 
 \includegraphics[width=\figwidthvar\linewidth]{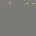}&
  \includegraphics[width=\figwidthvar\linewidth]{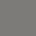} &
  \includegraphics[width=\figwidthvar\linewidth]{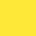} &
\includegraphics[width=\figwidthvar\linewidth]{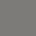} &
\includegraphics[width=\figwidthvar\linewidth]{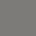}  \\
 \rotatebox{90}{Input}  & \rotatebox{90}{Ground-} \rotatebox{90}{Truth} &  \rotatebox{90}{Grad-} \rotatebox{90}{CAM} &  \rotatebox{90}{Grad-} \rotatebox{90}{CAMPP} &  \rotatebox{90}{Saliency} &  \rotatebox{90}{Deconv-} \rotatebox{90}{Net} &  \rotatebox{90}{Gradient-} \rotatebox{90}{Input} &  \rotatebox{90}{Guided-} \rotatebox{90}{Backprop}  &  \rotatebox{90}{Integrated-} \rotatebox{90}{Gradients}  &  \rotatebox{90}{Smooth-} \rotatebox{90}{Grad} & \rotatebox{90}{Square-} \rotatebox{90}{Grad}  &  \rotatebox{90}{VarGrad} & \rotatebox{90}{Occlusion} & \rotatebox{90}{Rise} & \rotatebox{90}{KernelShap} & \rotatebox{90}{Lime}\\
 \end{tabular}
    \caption{{Test set input images, ground truth and XAI method explanations.}}
    \label{fig:qualitative_results}
\end{figure*}

\begin{table*}[ht]
    \centering

    {    \scriptsize
\setlength{\tabcolsep}{0em} 
\begin{tabular}{lRRRRRRRRRRRRRRL@{\hskip 5pt}JJ}
{} &  \rotatebox{90}{Grad-} \rotatebox{90}{CAM} \rotatebox{90}{(2D)} &  \rotatebox{90}{Grad-} \rotatebox{90}{CAMPP} \rotatebox{90}{(2D)} &  \rotatebox{90}{Saliency} \rotatebox{90}{(2D)} &  \rotatebox{90}{Deconv-} \rotatebox{90}{Net (3D)} &  \rotatebox{90}{Gradient-} \rotatebox{90}{Input (3D)} &  \rotatebox{90}{Guided-} \rotatebox{90}{Backprop} \rotatebox{90}{(3D)}  &  \rotatebox{90}{Integrated-} \rotatebox{90}{Gradients} \rotatebox{90}{(3D)}  &  \rotatebox{90}{Smooth-} \rotatebox{90}{Grad (3D)} & \rotatebox{90}{Square-} \rotatebox{90}{Grad (3D)}  &  \rotatebox{90}{VarGrad} \rotatebox{90}{(3D)} & \rotatebox{90}{Occlusion} \rotatebox{90}{(2D)} & \rotatebox{90}{Rise (2D)} & \rotatebox{90}{KernelShap} \rotatebox{90}{(2D)} & \rotatebox{90}{Lime (2D)} & \rotatebox{90}{$\Delta$(min,max)} & \rotatebox{90}{Avg. metric} \rotatebox{90}{runtime}  & \rotatebox{90}{Avg. metric} \rotatebox{90}{runtime} \rotatebox{90}{(GPU)}\\
Del    &.18±0.0&.18±0.0&   .16±0.0 &    .13±0.0 &       .16±0.0 &        .13±0.0 &             .13±0.0 &     .13±0.0 &     .13±0.0 &     .13±0.0 &     .16±0.0 &        .13±0.0 &     .17±0.0 &     .16±0.0 &     .05           &  2.37±0.1     &   2.28±0.07    \\
Ins   &    .83±0.01 &    .83±0.01 &   .88±0.0 &    .94±0.0 &       .88±0.0 &        .95±0.0 &            .92±0.01 &     .93±0.0 &     .93±0.0 &    .93±0.01 &     .88±0.0 &        .94±0.0 &     .87±0.0 &    .88±0.01 &     .12   &   2.14±0.04    &    2.02±0.05   \\
MF  &  -.05±0.01 &  -.04±0.01 &  .01±0.02 &   .53±0.02 &       .7±0.02 &        .61±0.0 &            .76±0.01 &    .59±0.01 &    .66±0.01 &     .65±0.0 &    .01±0.01 &       .02±0.01 &     .0±0.01 &     .01±0.0 &     .81   &     30.28±0.13  &    21.69±0.16   \\
COS      &     .09±0.0 &     .09±0.0 &   .47±0.0 &    .61±0.0 &       .8±0.01 &        .69±0.0 &             .88±0.0 &     .77±0.0 &     .64±0.0 &     .62±0.0 &     .26±0.0 &        .08±0.0 &     .22±0.0 &     .21±0.0 &      .8   &    .04±0.04   &   .05±0.04    \\
SSIM        &     .4±0.01 &     .4±0.01 &   .87±0.0 &    .96±0.0 &       .98±0.0 &        .96±0.0 &             .98±0.0 &     .97±0.0 &     .96±0.0 &     .96±0.0 &     .83±0.0 &         .0±0.0 &     .41±0.0 &     .32±0.0 &     .98   &   .75±0.05    &    .88±0.04   \\
CSN &    .54±0.01 &    .54±0.01 &   .99±0.0 &    .99±0.0 &       1.0±0.0 &        .99±0.0 &             1.0±0.0 &     .99±0.0 &     .99±0.0 &     .99±0.0 &     .96±0.0 &         .0±0.0 &     .08±0.0 &     .09±0.0 &     1.0   &    .02±0.03   &   0.04±0.05   \\
F1          &     .03±0.0 &     .03±0.0 &  .55±0.01 &    .54±0.0 &      .76±0.01 &        .64±0.0 &            .85±0.01 &     .31±0.0 &     .3±0.01 &     .3±0.01 &     .19±0.0 &        .03±0.0 &     .02±0.0 &     .02±0.0 &     .83   &   1.37±0.05    &    1.56±0.05   \\
MAE         &     .22±0.0 &     .22±0.0 &   .01±0.0 &     .0±0.0 &        .0±0.0 &         .0±0.0 &              .0±0.0 &      .0±0.0 &      .0±0.0 &      .0±0.0 &     .03±0.0 &        .61±0.0 &     .17±0.0 &      .2±0.0 &      .6   &     .2±0.05   &   .23±0.05    \\
IoU         &     .02±0.0 &     .02±0.0 &   .42±0.0 &    .39±0.0 &      .72±0.01 &        .49±0.0 &            .82±0.01 &     .19±0.0 &     .18±0.0 &     .18±0.0 &     .11±0.0 &        .01±0.0 &     .01±0.0 &     .01±0.0 &     .81   &   1.27±0.06    &  1.45±0.04     \\
EBPG        &    .05±0.01 &    .05±0.01 &  .54±0.01 &    .44±0.0 &      .85±0.01 &        .57±0.0 &             .93±0.0 &     .53±0.0 &     .57±0.0 &    .51±0.01 &     .15±0.0 &        .02±0.0 &     .08±0.0 &     .06±0.0 &     .91   &    .07±0.05   &     .11±0.06  \\
\vspace{4pt}RRA         &     .4±0.02 &     .4±0.02 &   .99±0.0 &    .96±0.0 &       1.0±0.0 &        1.0±0.0 &             1.0±0.0 &    .89±0.01 &    .89±0.01 &    .88±0.01 &     .5±0.02 &        .2±0.02 &    .34±0.01 &    .35±0.01 &      .8   &   .06±0.04    &   .08±0.05    \\

cor!=       &     .33±0.0 &     .33±0.0 &   .63±0.0 &    .59±0.0 &      .86±0.01 &        .64±0.0 &             .93±0.0 &     .69±0.0 &     .74±0.0 &    .71±0.01 &    .43±0.01 &        .34±0.0 &     .45±0.0 &     .43±0.0 &     .59   &   .27±0.04    &   .31±0.03    \\
cor>        &     .33±0.0 &     .33±0.0 &   .56±0.0 &    .66±0.0 &      .85±0.01 &        .73±0.0 &             .92±0.0 &     .73±0.0 &     .65±0.0 &     .63±0.0 &     .43±0.0 &        .37±0.0 &     .43±0.0 &     .42±0.0 &     .59   &    .25±0.04   &    .3±0.04   \\
\vspace{4pt} cor<        &     .59±0.0 &     .59±0.0 &   .64±0.0 &    .49±0.0 &       .94±0.0 &        .52±0.0 &             .97±0.0 &    .61±0.01 &     .69±0.0 &     .68±0.0 &    .78±0.01 &        .51±0.0 &     .44±0.0 &    .39±0.01 &     .57   &   .27±0.04    &  .29±0.03     \\

cor$_{ns}$  &     .33±0.0 &     .33±0.0 &   .63±0.0 &    .59±0.0 &      .86±0.01 &        .64±0.0 &             .93±0.0 &     .69±0.0 &     .74±0.0 &    .71±0.01 &    .43±0.01 &        .34±0.0 &     .45±0.0 &     .43±0.0 &     .59   &    .25±0.04   &    .32±0.03   \\
cor$_{s}$    &     .46±0.0 &     .46±0.0 &    .6±0.0 &    .57±0.0 &        .9±0.0 &        .63±0.0 &             .94±0.0 &     .67±0.0 &     .67±0.0 &     .66±0.0 &     .61±0.0 &        .44±0.0 &     .44±0.0 &     .41±0.0 &     .54   &    .52±0.04   &   .61±0.03    \\
Time        &   6.3 &   6.1 &  4.1 &  6.5 &      .2 &      6.9 &          24.6 &  18.5 &  18.1 &  18.3 &  46.7 &  825.6 &  94.0 &  29.6 & 
\end{tabular}
}
 
      \caption{Evaluation of XAI methods using our and prior metrics.}
    \label{tab:xaimethod_results}
\end{table*}

\begin{table*}[ht]
    \centering

{    \scriptsize

\setlength{\tabcolsep}{0em} 
\begin{tabular}{lMMMMMMMMMMMMMML@{\hskip 5pt}JJ}
{} &  \rotatebox{90}{Grad-} \rotatebox{90}{CAM} \rotatebox{90}{(2D)} &  \rotatebox{90}{Grad-} \rotatebox{90}{CAMPP} \rotatebox{90}{(2D)} &  \rotatebox{90}{Saliency} \rotatebox{90}{(2D)} &  \rotatebox{90}{Deconv-} \rotatebox{90}{Net (3D)} &  \rotatebox{90}{Gradient-} \rotatebox{90}{Input (3D)} &  \rotatebox{90}{Guided-} \rotatebox{90}{Backprop} \rotatebox{90}{(3D)}  &  \rotatebox{90}{Integrated-} \rotatebox{90}{Gradients} \rotatebox{90}{(3D)}  &  \rotatebox{90}{Smooth-} \rotatebox{90}{Grad (3D)} & \rotatebox{90}{Square-} \rotatebox{90}{Grad (3D)}  &  \rotatebox{90}{VarGrad} \rotatebox{90}{(3D)} & \rotatebox{90}{Occlusion} \rotatebox{90}{(2D)} & \rotatebox{90}{Rise (2D)} & \rotatebox{90}{KernelShap} \rotatebox{90}{(2D)} & \rotatebox{90}{Lime (2D)} & \rotatebox{90}{$\Delta$(min,max)} & \rotatebox{90}{Avg. metric} \rotatebox{90}{runtime}  & \rotatebox{90}{Avg. metric} \rotatebox{90}{runtime} \rotatebox{90}{(GPU)}\\

PR          &    .04±0.01 &    .04±0.01 &  .49±0.01 &    .41±0.0 &      .85±0.01 &        .52±0.0 &             .93±0.0 &      .2±0.0 &     .2±0.01 &     .2±0.01 &     .14±0.0 &        .01±0.0 &     .03±0.0 &     .03±0.0 &     .91   &   1.33±0.07     &    1.52±0.05   \\
\vspace{4pt}RE          &    .82±0.01 &    .82±0.01 &  .72±0.01 &    .89±0.0 &      .72±0.01 &        .87±0.0 &            .82±0.01 &     .85±0.0 &    .85±0.01 &    .85±0.01 &    .55±0.01 &        1.0±0.0 &     .92±0.0 &     .91±0.0 &     .45   &    1.33±0.05   &   1.52±0.04    \\

cpa!=       &    .05±0.01 &    .05±0.01 &  .54±0.01 &    .44±0.0 &      .86±0.01 &        .56±0.0 &             .94±0.0 &      .6±0.0 &    .67±0.01 &     .6±0.01 &     .15±0.0 &        .01±0.0 &     .15±0.0 &      .1±0.0 &     .92   &   .17±0.05    &   .19±0.05    \\
cpl!=       &     .61±0.0 &     .61±0.0 &   .71±0.0 &    .73±0.0 &      .85±0.01 &        .72±0.0 &             .91±0.0 &     .77±0.0 &     .81±0.0 &     .81±0.0 &     .7±0.01 &        .66±0.0 &     .75±0.0 &     .77±0.0 &      .3   &    .11±0.04   &     .13±0.05  \\
cpa>        &    .04±0.01 &    .04±0.01 &   .4±0.01 &    .55±0.0 &      .86±0.01 &         .7$\pm$0.0 &             .94$\pm$0.0 &     .7$\pm$0.01 &    .51±0.01 &    .46±0.01 &     .14±0.0 &        .01±0.0 &      .1±0.0 &     .08±0.0 &     .93   &    .18±0.05   &    .2±0.05   \\
cpl>        &     .62±0.0 &     .62±0.0 &  .73±0.01 &    .77±0.0 &      .83±0.01 &        .76±0.0 &              .9±0.0 &     .77±0.0 &      .8±0.0 &     .81±0.0 &    .73±0.01 &        .72±0.0 &     .76±0.0 &     .75±0.0 &     .28   &    .09±0.05   &    .12±0.04   \\
cpa<        &     1.0±0.0 &     1.0±0.0 &   1.0±0.0 &    .37±0.0 &       1.0±0.0 &        .44$\pm$0.0 &             1.0$\pm$0.0 & .47$\pm$0.01 &     1.0±0.0 &     1.0±0.0 &     1.0±0.0 &        1.0±0.0 &    .24±0.01 &    .16±0.01 &     .84   &    .17±0.05   &     .2±0.05  \\
cpl<        &     .19±0.0 &     .19±0.0 &   .29±0.0 &    .61±0.0 &       .89±0.0 &        .61±0.0 &             .93±0.0 &     .75±0.0 &     .37±0.0 &     .36±0.0 &    .56±0.01 &        .02±0.0 &     .65±0.0 &    .63±0.01 &     .91   &   .1±0.05   &   .11±0.05    \\
 
\end{tabular}
}
      \caption{Evaluation of XAI methods using metrics that quantify fidelity in terms of precision or recall separately.}  \label{tab:xaimethod_results_2}
\end{table*}

We present the assessment of widely used XAI methods using our correctness metrics, as well as using widely used prior metrics in Table~\ref{tab:xaimethod_results}. Prior metrics are shown above the upper dividing white space, while our metrics are depicted below it. Our metrics are also divided into two. The metrics above the dividing white space are intended for detailed analysis, while Correctness$_{ns}$ and Correctness$_{s}$ aggregate these into single values. All values were obtained without a GPU accessible.  
Our metric suite contains compactness and completeness metrics that quantify the fidelity of explanations in terms of precision (i.e., compactness), or recall (i.e., completeness) separately. Also, it contains various versions of these metrics that evaluate XAI methods among negatively or positively contributing input nodes only. The average scores of these metrics are shown in Table~\ref{tab:xaimethod_results_2}. In Table~\ref{tab:xaimethod_results} and \ref{tab:xaimethod_results_2}, mean±std values over 3 runs are given. For all metrics except \textit{MSE}, \textit{Del} and \textit{Time} higher values are better. Explanation dimension is shown in brackets. Values are shaded. Runtime columns denote the average runtime in milliseconds to evaluate a single explanation with or without GPU accessible. Computational times of metrics that consult ground truth explanations do not include the time needed to construct the GT model and generate GT explanations.  Note that run-time differences between run-times with and without GPU accessible might be attributable to library version differences rather than possible GPU usage. 

To visualize explanations (Figure~\ref{fig:qualitative_results}), 3D explanations were converted to 2D. All explanations were normalized (see Section~\ref{sec:ground-truth}) and thereafter colormapped.

\section{{Discussion}}\label{sec:discussion}

Figure~\ref{fig:qualitative_results} shows test set input images, ground truth and XAI explanations. In row two, a concept that should be present at a specific location is present in other locations as well. All instances of this concept, irrespective of its location, are highlighted by GradCAM(PP). 
The dense sub-model filters out such irrelevant instances. GradCAM(PP) fails to explain this as their explanations are a weighted average of the last convolutional layer's output channels, i.e. of concept maps each containing all instances of a concept. 

In row two, an input example of a class that is defined by an XOR relation between two concepts is shown.  We argue two concepts should show up in ground truth images: the concept that is present and the concept that should not be present. Only DeconvNet, KernelShap and Lime (partially) adhere to this. 
DeconvNet back-propagates gradients through ReLu only if they are positive. Since $A~\text{XOR}~B \equiv (A \vee B) \wedge \neg(A \wedge B)$, DeconvNet includes $B$ in its explanation when $A=1$ and $B=0$, because it is in $A \vee B$ where it has positive gradient, but its role in $\neg(A \wedge B)$ is actually not explained as it has a negative gradient. Likewise, all concept parts of concepts with an 'OR' in their definition are included in explanations, even when they are not present in the input. \eg, concept 0 (Figure~\ref{fig:concept_parts}c) is always explained as the middle image. 

In row three, an input of class $4$ is shown. The input  belongs to this class because concept $2$ is not present at position $0-3$. We argue the explanations should show this. However, none of the methods completely adhere to this. 
XAI methods tend to not back-propagate gradients through a ReLu that is not activated in a forward pass, or not do so if the gradients themselves are negative. This makes them unable to explain that an input belongs to a certain class because a certain concept is not present. I.e., inactivated input nodes connected with negative weight to a node activated due to its positive bias tend to be excluded from explanations.

Assessment of widely used XAI methods using our evaluation approach, as well as prior metrics, is presented in Table~\ref{tab:xaimethod_results} and \ref{tab:xaimethod_results_2}. Metrics should assign a distinctly high value to good explanations and a low value to bad ones. The difference between the maximum and minimum value observed among XAI methods per metric is shown in the ``$\Delta$(min,max)'' column. For example, the difference between the maximum (1, see row 6 Table \ref{tab:xaimethod_results}) and minimum (0, see row 6) metric value of CSN is $1-0 = 1$. The compactness scores are among the most in value differing metrics. The Completeness$_<$ is among the most in value differing metrics as well, and the other completeness metrics have relatively low difference in value. Correctness distribution is the average of that of compactness and completeness.

Integrated gradients are generally judged as the best performing method. Note that making pixels darker (which is done by the method) ideally suits the explained method for classes where a concept must be present, while it does not for classes where a concept must not be present.

Our metric suite contains compactness and completeness metrics that quantify the fidelity of explanations in terms of precision (i.e., compactness), or recall (i.e., completeness) separately. Also, it contains various versions of these metrics that evaluate XAI methods among negatively or positively contributing input nodes only. See Table~\ref{tab:xaimethod_results_2}. Taking Smoothgrad as an example, we can observe from $cpa_> \approx 0.7, cpl_> \approx 0.77$ that the method has a good balance between recall and precision among positively contributing pixels. And, from $cpa_< \approx 0.47, cpl_< \approx 0.75$ we can observe that this is not the case among negatively contributing pixels. The method tends to explain non-negatively contributing input nodes as being negatively contributing. Obviously, the prior metrics precision and recall (top rows) do not provide these insights.

The mean time needed to evaluate an XAI metric on a single explanation is listed in Table~\ref{tab:xaimethod_results} and \ref{tab:xaimethod_results_2}. Our metrics, especially completeness and  compactness, are among the fastest metrics that were evaluated in our experiments. Correctness can be inferred directly from completeness and compactness, but all three were computed separately while taking time measurements. We used widely used implementations of prior metrics. Run-times of metrics (\eg, IoU, PR, RE, F1) might improve through code-level optimizations. 

Our metrics require ground truth explanations. Metrics that do not require  ground truth can be applied more broadly than those that do. In our work, input data and model realism were traded for ground truths being available. Our metrics require, in contrast to some prior metrics, normalized explanations and ground-truth explanations. Our normalization procedure could in theory cause problems in case an explanation contains very large or small outliers. 

Our evaluation approach may not assess all desirable properties of explanations. Metrics that measure the utility of explanations to humans without involving humans in the evaluation are \textit{proxy metrics}. \cite{rebuttual_1,rebuttual_2} claim that proxy metrics can (in some cases) align poorly with actual utility to humans. The XAI method evaluation approach we propose is fundamentally different from the ones discussed in said papers. Our GTs are derived from a synthetic model directly, unlike the discussed approaches that use human annotations as GTs. Also, the alternative to proxy metrics, i.e., involving humans in the evaluation of XAI methods, can be infeasible in practice and arguably introduces undesirable subjectivity. 

\section{{Conclusions}}\label{sec:conclusion}
The rationale behind a deep network's output is often difficult to understand.  EXplainable AI (XAI) methods try to solve this problem by explicate certain predictions by a given model. Reliable evaluation metrics are required to assess and compare different XAI methods. We proposed a novel evaluation approach for benchmarking XAI attribution methods. Our proposal consists of a synthetic classification model accompanied by its derived ground truth explanations.  We also proposed novel, high-fidelity metrics to quantify the difference between explanations of the XAI method under investigation and those derived from the synthetic model. Our metrics allow accurate assessment of explanations in terms of precision and recall separately and can independently evaluate negative or positive contributions of input nodes. We investigated our evaluation approach by benchmarking widely used XAI attribution methods and compared against established prior XAI evaluation metrics. Our proposal provided deeper insights into XAI methods output. For example, Guided-Backprop and Smoothgrad show good precision and recall scores among positively contributing pixels, but poor precision scores among negatively contributing pixels. The recall scores in the latter case remain close.  Moreover, our metrics are among the fastest in terms of execution time.

\subsubsection*{Acknowledgements}

Financial support by the Groningen Cognitive Systems and Materials Center (CogniGron) and the Ubbo Emmius Foundation of the University of Groningen is gratefully acknowledged.

{\small
\bibliographystyle{ieee_fullname}
\bibliography{main}
}

\end{document}